# Minimax regret based elicitation of generalized additive utilities

**Darius Braziunas**
Department of Computer Science
University of Toronto
Toronto, ON M5S 3G4
*darius@cs.toronto.edu*

**Craig Boutilier**
Department of Computer Science
University of Toronto
Toronto, ON M5S 3G4
*cebly@cs.toronto.edu*

## Abstract

We describe the semantic foundations for elicitation of generalized additively independent (GAI) utilities using the minimax regret criterion, and propose several new query types and strategies for this purpose. Computational feasibility is obtained by exploiting the local GAI structure in the model. Our results provide a practical approach for implementing preference-based constrained configuration optimization as well as effective search in multiattribute product databases.

## 1 Introduction

Representing, reasoning with, and eliciting the preferences of individual users is a fundamental problem in the design of decision support tools (and indeed, much of AI). A key issue in preference research is dealing with large, multi-attribute problems: preference representation and elicitation techniques must cope with the exponential size of the outcome space. By exploiting intrinsic independencies in preference structure, *factored* utility models can provide tractable (although sometimes approximate) solutions to the problem. Utility function structure (such as additive, multilinear, generalized additive, etc.) can be used to represent large utility models very concisely [10]. While additive models are the most widely used in practice, *generalized additive independence models* (GAI) have recently generated interest because of their greater flexibility and applicability [1, 3, 8, 4, 6]. Even though the semantic foundations of the GAI representation were described by Fishburn decades ago [7], the design of effective elicitation techniques has gained attention only recently [4, 8, 6].

In this paper, we develop a new model for utility elicitation in GAI models based on the minimax regret decision criterion [13, 11]. Minimax regret provides a robust way of making decisions under utility function uncertainty, minimizing worst-case loss under all possible realizations of a user's utility function [3, 12, 4]; as such it is applicable when distributional information over utility functions is not easily available. Regret has also proven to be an effective driver of preference elicitation [14, 4, 5]. However, prior work on regret-based elicitation for GAI models has ignored key semantic issues, thus simplifying the approach to both elicitation and regret computation and optimization. By building on the semantic foundations of GAI elicitation laid out in [6], we identify new classes of elicitation queries suitable for regret-based elicitation, and propose several new query strategies based on these classes.

Our approach emphasizes *local queries* over small sets of attributes; but *global queries* over full outcomes are required to calibrate certain terms across GAI factors (a problem ignored in previous work on regret-based elicitation). However, we will demonstrate that most of the "heavy lifting" can be achieved using local queries. Our new approach guarantees the semantic soundness of the utility representation in a way that techniques that ignore interactions across factors do not. In addition, our new queries impose much more intricate constraints on GAI model parameters than those considered in previous work. For this reason, we develop new formulations of the linear mixed integer programs (MIPs) that are used in regret-based optimization, and show that the problem can be effectively solved despite the added complexity.

We begin in Sec. 2 with relevant background on multiattribute utility. We discuss appropriate forms of both local and global queries for GAI elicitation in Sec. 3. We then describe effective MIP formulations for minimax regret computation in Sec. 4, including discussion of regret computation in multiattribute product databases. Sec. 5 presents several elicitation strategies based on the query types above as well as empirical evaluation. Future directions are summarized in Sec. 6.

## 2 Multiattribute preferences

Assume a set of attributes $X_1, X_2, \ldots, X_n$, each with finite domains, which define a set of *outcomes* $\mathbf{X} = X_1 \times \cdots \times X_n$. The preferences of a user, on whose behalf decisions are made, are captured by a *utility function* $u : \mathbf{X} \mapsto \mathbb{R}$. A utility function can viewed as reflecting (qualitative) pref-erences over *lotteries* (distributions over outcomes) [10], with one lottery preferred to another iff its expected utility is greater. Let $\langle p, \mathbf{x}^\top; 1-p, \mathbf{x}^\perp \rangle$ denote the lottery where the best outcome $\mathbf{x}^\top$ is realized with probability $p$, and the worst outcome $\mathbf{x}^\perp$ with probability $1-p$; we refer to best and worst outcomes as *anchor* outcomes. Since utility functions are unique up to positive affine transformations, it is customary to set the utility of the best outcome $\mathbf{x}^\top$ to 1, and the utility of the worst outcome $\mathbf{x}^\perp$ to 0. If a user is indifferent between some outcome $\mathbf{x}$ and the *standard gamble* $\langle p, \mathbf{x}^\top; 1-p, \mathbf{x}^\perp \rangle$, then $u(\mathbf{x}) = p$.

## 2.1 Additive utilities

Since the size of outcome space is exponential in the number of attributes, specifying the utility of each outcome is infeasible in many practical applications. Most preferences, however, exhibit internal structure that can be used to express $u$ concisely. *Additive independence* [10] is commonly assumed in practice, where $u$ can be written as a sum of single-attribute *subutility functions*:[1]

$$u(\mathbf{x}) = \sum_{i=1}^n u_i(x_i) = \sum_{i=1}^n \lambda_i v_i(x_i).$$

The subutility functions $u_i(x_i) = \lambda_i v_i(x_i)$ can be defined as a product of *local value functions (LVFs)* $v_i$ and scaling constants $\lambda_i$. This simple factorization allows us to separate the representation of preferences into two components: "local" and "global." Significantly, LVFs can be defined using "local" lotteries that involve only a single attribute: $v_i(x_i) = p$, where $p$ is the probability at which the user is indifferent between two local outcomes $x_i$ and $\langle p, x_i^\top; 1-p, x_i^\perp \rangle$, *ceteris paribus*.[2] Since we can define value functions independently of other attributes, we can also *assess* them independently using queries only about values of attribute $i$. This focus on preferences over individual attributes has tremendous practical significance, because people have difficulty taking into account more than five or six attributes at a time [9].

The scaling constants $\lambda_i$ are "global" and are required to properly calibrate LVFs across attributes. To define the scaling constants, we first introduce a notion of a *reference* (or *default*) outcome, denoted by $\mathbf{x}^0 = (x_1^0, x_2^0, \ldots, x_n^0)$. The reference outcome is an arbitrary outcome, though it is common to choose the worst outcome $\mathbf{x}^\perp$ as $\mathbf{x}^0$ (more generally, any *salient* outcome, such as an "incumbent" will prove useful in this role). Let $\mathbf{x}^{\top i}$ be a full outcome where the $i^{th}$ attribute is set to its best level whereas other attributes are fixed at their reference levels; $\mathbf{x}^{\perp i}$ is defined similarly. Then, $\lambda_i = u(\mathbf{x}^{\top i}) - u(\mathbf{x}^{\perp i})$. To assess scaling constants $\lambda_i$, one must ask queries about $2n$ global outcomes $\mathbf{x}^{\top i}$ and $\mathbf{x}^{\perp i}$ for each attribute $i$.[3] These global outcomes are special because they involve varying only a single feature from the reference outcome. This ease of assessment makes additive utility the model of choice in most practical applications.

## 2.2 Generalized additive utilities

Simple additive models, although very popular in practice, are quite restrictive in their assumptions of attribute independence. A more general utility decomposition, based on *generalized additive independence* (GAI), has recently gained more attention because of its additional flexibility [1, 3, 8, 4, 6]. It can model "flat" utility functions with no internal structure as well as linear additive models. Most realistic problems arguably fall somewhere between these two extremes.

GAI models [7, 1] additively decompose a utility function over (possibly overlapping) *subsets* of attributes. Formally, assume a given collection $\{I_1, \ldots, I_m\}$ of possibly intersecting attribute (index) sets, or *factors*. Given an index set $I \subseteq \{1, \ldots, n\}$, we define $\mathbf{X}_I = \times_{i \in I} X_i$ to be the set of *partial outcomes* (or *suboutcomes*) restricted to attributes in $I$. For a factor $j$, $\mathbf{x}_{I_j}$, or simply $\mathbf{x}_j$, is a particular instantiation of attributes in factor $j$. The factors are *generalized additively independent* if and only if the user is indifferent between any two lotteries with the same marginals on each set of attributes [7]. Furthermore, if GAI holds, the utility function can be written as a sum of *subutility* functions [7]:

$$u(\mathbf{x}) = u_1(\mathbf{x}_{I_1}) + \ldots + u_m(\mathbf{x}_{I_m}).$$

The key difference between additive and GAI models with regard to elicitation (rather than representation) lies in the *semantics* of subutility functions $u_i$. In additive models, the quantities $u_i(x_i) = \lambda_i v_i(x_i)$ have a very clear decision-theoretic meaning.[4] In contrast, GAI subutility functions *are not unique* and, in the absence of further qualifications, do not have a well-defined semantic interpretation. This makes elicitation of GAI model parameters problematic. Intuitively, since utility can "flow" from one subutility factor to another through shared attributes, the values of subutility $u_j$ do not directly represent the local preference relation among the attributes in factor $j$.

For effective elicitation we therefore need a representation of GAI utilities such that: 1) all GAI model parameters have a sound semantic interpretation; and, 2) the GAI structure is reflected by separating the parameters into *local* and *global* groups, in a way analogous to additive models. Building on the foundational work of Fishburn [7], we [6]

---

[1] This decomposition is possible iff a user is indifferent between lotteries with the same marginals on each attribute.

[2] $x_i^\top$ and $x_i^\perp$ are the best and worst levels of attribute $i$. Without loss of generality, we assume $v_i(x_i^\top) = 1$, $v_i(x_i^\perp) = 0$.

[3] Only $n$ outcomes if the reference outcome is also the worst.

[4] In additive utility models, the LVF $v_i(x_i)$ is simply the probability $p$ at which the user is indifferent between $x_i$ and $\langle p, x_i^\top; 1-p, x_i^\perp \rangle$, *ceteris paribus*, and $\lambda_i$ is $u(\mathbf{x}^{\top i}) - u(\mathbf{x}^{\perp i})$.

demonstrate that the following *canonical* representation of GAI utilities achieves both goals:

$$u(\mathbf{x}) = \sum_{j=1}^{m} u_j(\mathbf{x}_j) = \sum_{j=1}^{m} \lambda_j \, \bar{u}_j(\mathbf{x}_j). \quad (1)$$

Here, similar to additive models, $\lambda_j$ is a scaling constant, and $\bar{u}_j$ is an *unscaled* subutility function, which itself is a sum of the values of a function $v_j$ (to be defined later) at certain suboutcomes related to $\mathbf{x}_j$:

$$\bar{u}_j(\mathbf{x}_j) = v_j(\mathbf{x}_j) + \sum_{k=1}^{j-1}(-1)^k \sum_{1 \leq i_1 < \cdots i_k < j} v_j(\mathbf{x}_j[\bigcap_{s=1}^{k} I_{i_s} \cap I_j]). \quad (2)$$

The sum in the equation is only over non-empty intersections $\bigcap_{s=1}^{k} I_{i_s} \cap I_j$. For any $\mathbf{x}$, $\mathbf{x}[I]$ is an outcome where attributes not in $I$ are set to the reference value (i.e., $X_i = x_i$ if $i \in I$, and $X_i = x_i^0$ if $i \notin I$). For further details, we refer to [6].

Our key result [6] shows that the function $v_j$ in Eq. 2 generalizes LVFs defined earlier for additive models. Let the *conditioning set* $K_j$ of factor $j$ be the set of all attributes that share GAI factors with attributes in $j$. Intuitively, fixing the attributes in the conditioning set to any value "blocks" the influence of other factors on factor $j$. In a manner similar to additive models, the local value $v_j(\mathbf{x}_j)$ of suboutcome $\mathbf{x}_j$ is simply $p$, the probability that induces indifference between $\mathbf{x}_j$ and the *local* standard gamble $\langle p, \mathbf{x}_j^\top; 1-p, \mathbf{x}_j^\bot \rangle$, *given that attributes in the conditioning set $K_j$ are fixed at reference levels*, *ceteris paribus*. We refer to the setting of attributes in $K_j$ to their reference values (ceteris paribus) as the *local value condition*. Here $\mathbf{x}_j^\top$ and $\mathbf{x}_j^\bot$ are the best and worst suboutcomes in factor $j$ assuming the local value condition; by definition, the LVFs are normalized, so $v_j(\mathbf{x}_j^\top) = 1$ and $v_j(\mathbf{x}_j^\bot) = 0$. We see, then, that LVFs have a very clear semantic interpretation; they calibrate local preferences relative to the best and worst factor suboutcomes under the local value condition. Thus LVFs are *local*, involving only attributes in single factors and their (usually small) conditioning sets.

The global scaling constants $\lambda_j$ are defined in a way analogous to the additive utility case. Let $\mathbf{x}^{\top j}$ and $\mathbf{x}^{\bot j}$ be the best and the worst (full) outcomes, given that attributes not in factor $j$ are set to their reference levels. Then, $\lambda_j = u(\mathbf{x}^{\top j}) - u(\mathbf{x}^{\bot j}) = u_j^\top - u_j^\bot$. We will refer to $u_j^\top$ and $u_j^\bot$ as *anchor utilities* for factor $j$.

To compute the unscaled subutility function $\bar{u}_j(\mathbf{x}_j)$, one needs to know which local suboutcomes are involved (in the form $\mathbf{x}_j[\bigcap_{s=1}^{k} I_{i_s} \cap I_j]$) on the right-hand side of Eq. 2; this amounts to finding all nonempty sets $\bigcap_{s=1}^{k} I_{i_s} \cap I_j$ and recording the sign $(+/-)$ for the corresponding LVFs. We refer to this procedure as computing the *dependency structure* of a GAI model. An efficient graphical search algorithm for computing such dependencies among LVFs was first described in [6].

Knowing the dependency structure, Eq. 2 can be simplified by introducing the following notation. Let $N_j$ be the number of local configurations (settings of attributes) in factor $j$ (e.g., with 3 boolean attributes, $N_j = 8$). The LVF $v_j$ can be expressed by $N_j$ parameters $v_j^1, \ldots, v_j^{N_j}$ such that $v_j^i = v_j(\mathbf{x}_j)$, where $i$ is the index of the local configuration $\mathbf{x}_j$. Then, Eq. 2 can be rewritten as

$$\bar{u}_j(\mathbf{x}_j) = \sum_{i \in 1..N_j} C_{\mathbf{x}_j}^i \, v_j^i, \quad (3)$$

where the $C_{\mathbf{x}_j}^i$ are integer coefficients precomputed using the dependency structure (most of these are zero).

Thus, a GAI model, similar to simple additive utility functions, is additively decomposed into factors that are a product of scaling constants, or *weights*, $\lambda_j = u_j^\top - u_j^\bot$, and a linear combination of LVF parameters:

$$u(\mathbf{x}) = \sum_j \left[ (u_j^\top - u_j^\bot) \sum_{i \in 1..N_j} C_{\mathbf{x}_j}^i \, v_j^i \right]. \quad (4)$$

This representation of GAI utilities achieves the goals described above: the representation is unique, all parameters have a well-defined semantics, and they are grouped into local (LVFs) and global (anchor utilities) parameters. The next section introduces appropriate queries for assessing these GAI model parameters. (The GAI model *structure* is represented by parameters $C_{\mathbf{x}_j}^i$.)

## 3 Elicitation queries

In general, eliciting complete preference information is costly and, in most cases, unnecessary for making an optimal decision. Instead, elicitation and decision making can be viewed as a single process whose goal is to achieve the right tradeoff between the costs of interaction, potential improvements of decision quality due to additional preference information, and the value of a terminal decision [2].

The types of queries one considers is an integral part of the preference elicitation problem. Some queries are easy to answer, but do not provide much information, while more informative queries are often costly. Computing or estimating the value of information can vary considerably for different query types. Finally, allowable queries define the nature of constraints on the feasible utility set. We broadly distinguish *global queries* over full outcomes from *local queries* over subsets of outcomes. In most multiattribute problems, people can meaningfully compare outcomes with no more than five or six attributes [9]. Therefore, we propose local counterparts to global queries that apply to a subset of attributes.

From Eq. 4, we can see that a GAI utility function can be fully assessed by eliciting the LVF parameters $v_j^i$ and the anchor utilities $u_j^\top$ and $u_j^\bot$. The LVF parameters can be determined by posing *local queries*; such queries do not require a user to consider the values of all attributes. In addition, to achieve the right calibration of the LVFs, we need to elicit utilities of a few *full* outcomes: for each factor, we must know the utility of the best and the worst outcomes given that attributes in *other* factors are set to their reference levels (i.e., elicit the values $u_j^\top$ and $u_j^\perp$).

We will consider four types of queries for elicitation. The following queries are well-defined semantically, relatively simple, and easy to explain to non-expert users.

**Local bound queries** An LVF calibrates utilities of partial outcomes with respect to partial "anchor" outcomes $\mathbf{x}_j^\top$ and $\mathbf{x}_j^\perp$, given that the attributes in conditioning set $K_j$ are fixed at their reference levels. A *local bound (LB)* query on parameter $v_j^\mathbf{x}$ is as follows: "Assume that the attributes in $K_j$ are fixed at reference levels. Would you prefer the partial outcome $\mathbf{x}_j$ to a lottery $\langle \mathbf{x}_j^\top, p; \mathbf{x}_j^\perp, 1-p \rangle$, assuming that the remaining attributes are fixed at same levels (ceteris paribus)?" If the answer is "yes", $v_j^\mathbf{x} \geq p$; if "no", then $v_j^\mathbf{x} < p$. By definition, the local value parameters $v_j^\mathbf{x}$ lie in [0,1]. This binary (yes/no) query differs from a direct (local) standard gamble query since we do not ask the user to *choose* the indifference level $p$, only bound it.

**Local comparison queries** Local comparisons are natural and easy to answer. A *local comparison (LC)* query asks a user to compare two partial outcomes: "Assume that the attributes in $K_j$ are fixed at reference levels. Would you prefer partial outcome $\mathbf{x}_j$ to partial outcome $\mathbf{x}_j'$, ceteris paribus?" If the answer is "yes", $v_j^\mathbf{x} \geq v_j^{\mathbf{x}'}$; if "no", then $v_j^\mathbf{x} < v_j^{\mathbf{x}'}$.

**Anchor bound queries** The scaling constant, or weight, for a subutility function $\bar{u}_j$ is $\lambda_j = u_j^\top - u_j^\perp$, where $u_j^\top$ is the global utility of the outcome in which the $j^{th}$ factor is set to its best value, and all the other attributes are fixed at reference levels. Similarly, $u_j^\perp$ is the utility of the "bottom anchor" of factor $j$. Utilities of anchor levels $u_1^\top, u_1^\perp, \ldots, u_m^\top, u_m^\perp$ must be obtained using global queries. However, we need only ask $2m$ direct utility queries over full outcomes; this is the *same number of global queries* required for scaling in the additive case (considering each attribute as a factor).

Instead of eliciting exact anchor utilities directly, we propose global queries that are easier to answer. An *anchor bound (AB)* query asks: "Consider a full outcome $\mathbf{x}^{\top j}$, where attributes in factor $j$ are set to their best values, and other attributes are fixed at reference levels. Do you prefer $\mathbf{x}^{\top j}$ to a lottery $\langle \mathbf{x}^\top, p; \mathbf{x}^\perp, 1-p \rangle$?" A "yes" response gives $u_j^\top \geq p$; and "no", $u_j^\top < p$ (assuming, without loss of generality, that $u(\mathbf{x}^\top) = 1$ and $u(\mathbf{x}^\perp) = 0$). An analogous query exists for the "bottom" anchor $\mathbf{x}^{\perp j}$.

**Anchor comparison queries** We can also ask a user to compare anchor outcomes from different factors: "Do you prefer global outcome $\mathbf{x}^{\top k}$ to $\mathbf{x}^{\perp l}$?" If "yes", then $u_k^\top \geq u_l^\perp$; if "no", then $u_k^\top < u_l^\perp$. Such *anchor comparison (AC)* queries are usually much easier to answer than anchor bound queries.

## 4 Minimax regret calculation

In our model, the uncertainty over user utility functions is defined by (linear) constraints on utility function parameters, specifically, those induced by responses to queries of the form above. Without distributional information w.r.t. possible utility functions, the *minimax regret* decision criterion is especially suitable. It requires that we recommend a feasible outcome $\mathbf{x}^*$ that minimizes *maximum regret* with respect to all possible realizations of the user's utility function [3, 12, 4]. This guarantees worst-case bounds on the quality of the decision made under the type of strict uncertainty induced by the queries above [14, 4, 5]. In case further preference information is available, a regret-based elicitation policy can be employed to reduce utility uncertainty and minimize interaction costs to the extent where an (approximately) optimal decision can be recommended (see Sec. 5).

Let $\mathbf{U}$ be the set of feasible utility functions, defined by constraints—induced by user responses to queries—on the values of factor anchors $u_j^\top, u_j^\perp$ (for each factor $j$), and constraints on the LVF parameters $v_j^i$. Let $Feas(\mathbf{X}) \subseteq \mathbf{X}$ be the set of *feasible* outcomes (e.g., defined by a set of hard constraints $\mathcal{H}$). We define minimax regret in three stages (following [4]). The *pairwise regret* of choosing $\mathbf{x}$ instead of $\mathbf{x}'$ w.r.t. $\mathbf{U}$ is $R(\mathbf{x}, \mathbf{x}', \mathbf{U}) = \max_{u \in \mathbf{U}} u(\mathbf{x}') - u(\mathbf{x})$. The *maximum regret* of choosing outcome $\mathbf{x}$ is $MR(\mathbf{x}, \mathbf{U}) = \max_{\mathbf{x}' \in Feas(\mathbf{X})} R(\mathbf{x}, \mathbf{x}', \mathbf{U})$. Finally, the outcome that minimizes max regret is the *minimax optimal decision*: $MMR(\mathbf{U}) = \min_{\mathbf{x} \in Feas(\mathbf{X})} MR(\mathbf{x}, \mathbf{U})$. We develop tractable formulations of these definitions for GAI models.

**Pairwise regret** Given a GAI model, the *pairwise regret* of $\mathbf{x}$ w.r.t. $\mathbf{x}'$ over $\mathbf{U}$ is:

$$R(\mathbf{x}, \mathbf{x}', \mathbf{U}) = \max_{u \in \mathbf{U}} u(\mathbf{x}') - u(\mathbf{x}) \qquad (5)$$

$$= \max_{u \in \mathbf{U}} \sum_j [u_j(\mathbf{x}_j') - u_j(\mathbf{x}_j)]$$

$$= \max_{\{u_j^\top, u_j^\perp, v_j^i\}} \sum_j (u_j^\top - u_j^\perp)(\bar{u}_j(\mathbf{x}_j') - \bar{u}_j(\mathbf{x}_j))$$

$$= \max_{\{u_j^\top, u_j^\perp, v_j^i\}} \sum_j \left[ (u_j^\top - u_j^\perp) \sum_{i \in 1..N_j} (C_{\mathbf{x}_j'}^i - C_{\mathbf{x}_j}^i) v_j^i \right].$$

In general, when constraints on utility space tie together parameters from different factors, regret computation has a quadratic objective. Such constraints might arise, for example, from global comparison queries. With linear constraints, this becomes a quadratic program.

Since factors reflect intrinsic independencies among attributes, it is natural to assume that utility constraints involve only parameters *within the same factor*. The constraints induced by *local* comparison or bound queries, for instance, have this form. We call constraints involving parameters within a single factor *local*. This allows modeling regret computation linearly as we discuss below.

If the constraints on local value parameters $v_j^i$ are local then Eq. 5 can be simplified by pushing one "max" inward. This is made possible by the fact that the scaling factors $u_j^\top - u_j^\perp$ are *always positive*: $R(\mathbf{x}, \mathbf{x}', \mathbf{U}) =$

$$= \max_{\{u_j^\top, u_j^\perp, v_j^i\}} \sum_j \left[ (u_j^\top - u_j^\perp) \sum_{i \in 1..N_j} (C_{\mathbf{x}'_j}^i - C_{\mathbf{x}_j}^i) v_j^i \right]$$

$$= \max_{\{u_j^\top, u_j^\perp\}} \sum_j \left[ (u_j^\top - u_j^\perp) \max_{\{v_j^i\}} \sum_{i \in 1..N_j} (C_{\mathbf{x}'_j}^i - C_{\mathbf{x}_j}^i) v_j^i \right]$$

$$= \max_{\{u_j^\top, u_j^\perp\}} \sum_j (u_j^\top - u_j^\perp) \bar{r}_j^{\mathbf{x}, \mathbf{x}'}, \qquad (6)$$

where (unscaled) "local regret"

$$\bar{r}_j^{\mathbf{x}, \mathbf{x}'} = \max_{\{v_j^i\}} \sum_{i \in 1..N_j} (C_{\mathbf{x}'_j}^i - C_{\mathbf{x}_j}^i) v_j^i \qquad (7)$$

can be precomputed by solving a small linear program (whose size is bounded by the factor size).

If constraints on LVF parameters are bound constraints only, and therefore independent of each other, we can do without linear programming when computing the local regret $\bar{r}_j^{\mathbf{x}, \mathbf{x}'}$ (by pushing the max within the sum):

$$\bar{r}_j^{\mathbf{x}, \mathbf{x}'} = \sum_{i \in 1..N_j} \max_{\{v_j^i\}} (C_{\mathbf{x}'_j}^i - C_{\mathbf{x}_j}^i) v_j^i,$$

where $\max_{\{v_j^i\}} (C_{\mathbf{x}'_j}^i - C_{\mathbf{x}_j}^i) v_j^i =$

$$\begin{cases} (C_{\mathbf{x}'_j}^i - C_{\mathbf{x}_j}^i) \max(v_j^i), & \text{if } C_{\mathbf{x}'_j}^i - C_{\mathbf{x}_j}^i \geq 0, \\ (C_{\mathbf{x}'_j}^i - C_{\mathbf{x}_j}^i) \min(v_j^i), & \text{if } C_{\mathbf{x}'_j}^i - C_{\mathbf{x}_j}^i < 0. \end{cases}$$

**Maximum regret** The *max regret* of choosing $\mathbf{x}$ is $MR(\mathbf{x}, \mathbf{U})$:

$$= \max_{\mathbf{x}' \in Feas(\mathbf{X})} R(\mathbf{x}, \mathbf{x}', \mathbf{U}) \qquad (8)$$

$$= \max_{\mathbf{x}' \in Feas(\mathbf{X}), u \in \mathbf{U}} u(\mathbf{x}') - u(\mathbf{x})$$

$$= \max_{\mathbf{x}' \in Feas(\mathbf{X}), \{u_j^\top, u_j^\perp, v_j^i\}} \sum_j \left[ (u_j^\top - u_j^\perp) \sum_{i \in 1..N_j} (C_{\mathbf{x}'_j}^i - C_{\mathbf{x}_j}^i) v_j^i \right]$$

If local value constraints involve only local value parameters within their own factors, the max regret expression above simplifies to:

$$MR(\mathbf{x}, \mathbf{U}) = \max_{\mathbf{x}' \in Feas(\mathbf{X}), \{u_j^\top, u_j^\perp\}} \sum_j (u_j^\top - u_j^\perp) \bar{r}_j^{\mathbf{x}, \mathbf{x}'}, \qquad (9)$$

where local regrets $\bar{r}_j^{\mathbf{x}, \mathbf{x}'}$ can be precomputed beforehand and treated as constants.

This optimization can be recast as a quadratic MIP:

$$MR(\mathbf{x}, \mathbf{U}) = \max_{\mathbf{x}' \in Feas(\mathbf{X})} \max_{\{u_j^\top, u_j^\perp\}} \sum_j (u_j^\top - u_j^\perp) \bar{r}_j^{\mathbf{x}, \mathbf{x}'}$$

$$= \max_{\{Z_j^{\mathbf{x}'}, u_j^\top, u_j^\perp\}} \sum_j \sum_{\mathbf{x}'_j} (u_j^\top - u_j^\perp) \bar{r}_j^{\mathbf{x}, \mathbf{x}'} Z_j^{\mathbf{x}'},$$

subject to constraints $\mathcal{A}, \mathcal{H}$ and $\mathcal{U}$,

where $\mathcal{A}$ are state definition constraints tying binary indicators $Z_j^{\mathbf{x}'}$ with consistent attribute assignments, and $\mathcal{H}$ are domain constraints defining feasible configurations. For each factor $j$, only one of the indicators $Z_j^{\mathbf{x}'} = 1$.

Using the "big-M" transformation, the quadratic optimization above can be linearized by introducing variables $Y_j^{\mathbf{x}'}$ which can be thought of as representing the product $(u_j^\top - u_j^\perp) Z_j^{\mathbf{x}'}$:[5]

$$MR(\mathbf{x}, \mathbf{U}) = \max_{\{Y_j^{\mathbf{x}'}, Z_j^{\mathbf{x}'}, u_j^\top, u_j^\perp\}} \sum_j \sum_{\mathbf{x}'_j} \bar{r}_j^{\mathbf{x}, \mathbf{x}'} Y_j^{\mathbf{x}'}, \qquad (10)$$

$$\text{subject to} \begin{cases} 0 \leq Y_j^{\mathbf{x}'} \leq M_j Z_j^{\mathbf{x}'}, \ \forall j, \mathbf{x}'_j, \\ \sum_{\mathbf{x}'_j} Y_j^{\mathbf{x}'} = u_j^\top - u_j^\perp, \ \forall j, \\ \mathcal{A}, \mathcal{H} \text{ and } \mathcal{U}. \end{cases}$$

In the formulation above, the first constraint ensures that $Y_j^{\mathbf{x}'} = 0$ whenever $Z_j^{\mathbf{x}'} = 0$. If $Z_j^{\mathbf{x}'}$ is one, $Y_j^{\mathbf{x}'}$ is bounded by some constant $M_j \geq u_j^\top - u_j^\perp$, and the second constraint ensures that $Y_j^{\mathbf{x}'}$ achieves the optimal value of $u_j^\top - u_j^\perp$. The $Y_j^{\mathbf{x}'} \geq 0$ constraint is included because the difference $u_j^\top - u_j^\perp$ is by definition always positive. Since the objective is now linear, the problem is a linear MIP.

**Minimax regret** Our goal is to find a feasible configuration $\mathbf{x}^*$ that minimizes maximum regret

$$MMR(\mathbf{U}) = \min_{\mathbf{x} \in Feas(\mathbf{X})} MR(\mathbf{x}, \mathbf{U}).$$

We can express this optimization as (linear) MIP, too:

$$MMR(\mathcal{U}) = \min_{\{Z_j^{\mathbf{x}}, m\}} m, \text{ subject to} \qquad (11)$$

$$\begin{cases} m \geq \sum_{j, \mathbf{x}_j} [(u_j^\top - u_j^\perp) \bar{r}_j^{\mathbf{x}, \mathbf{x}'}] Z_j^{\mathbf{x}}, \ \forall \mathbf{x}' \in Feas(\mathbf{X}), u_j^\top, u_j^\perp \in \mathbf{U}^v, \\ \mathcal{A}, \mathcal{H} \text{ and } \mathcal{U}, \end{cases}$$

where $\mathbf{U}^v \subset \mathbf{U}$ is a set of vertices of the polytope that defines the feasible values of anchor utilities $u_j^\top, u_j^\perp$. In practice, we avoid the exponential number of constraints (one for each feasible adversary configuration $\mathbf{x}'$ and anchor utilities $u_j^\top, u_j^\perp$) using an iterative constraint generation procedure that generates the (small) set of active constraints at the optimal solution. This requires solving the MIP in Eq. 11 with only a subset of constraints, generating the maximally violated constraint at the solution of this relaxed MIP (by solving the max regret MIP in Eq. 10 for the factor regrets $(u_j^\top - u_j^\perp)\bar{r}_j^{\mathbf{x},\mathbf{x}'}$), and repeating until no violated constraints are found (see [4] for details).

---

[5] In practice, we need not introduce extra variables and constraints, instead placing attribute consistency constraints $\mathcal{A}$ directly on continuous variables $Y_j^{\mathbf{x}'}$. However, the somewhat more transparent formulation here is presented for clarity (and does not perform significantly worse computationally).

**Multiattribute product databases** The MIP formulations above assume that the space of feasible configurations is defined by a set of constraints $\mathcal{H}$ specifying allowable combinations of attributes. Alternatively, the set of choices may be the elements of a *multiattribute product database*, in which the set of feasible outcomes is specified explicitly, namely, as the set of all products in the database. Preference-based search of, and choice from, such a database can be effected using minimax regret as well, but can in fact be somewhat simpler computationally.

For any two database items $\mathbf{x}$ and $\mathbf{x}'$, pairwise regret $R(\mathbf{x}, \mathbf{x}', \mathbf{U})$ can be computed using Eq. 6. The max regret $MR(\mathbf{x})$ of $\mathbf{x}$ is determined by considering its pairwise regret with each other item. To determine the optimal product (i.e., with minimax regret), we compute the $MR(\mathbf{x})$ of each item $\mathbf{x}$ and choose the one with least max regret. This latter computation can be sped up considerably by iteratively generating minimax optimal candidate products against a current set of "adversary" items and testing their optimality. In practice, much like constraint generation, this speed up reduces the complexity of the algorithm from quadratic to linear in the size of the database.

## 5 Elicitation strategies

Minimax regret allows one to bound the loss associated with the recommended decision relative to the (unknown) optimal. If this bound on utility loss is too high, more utility information must be elicited. A decision support system can query the user until minimax regret reaches some acceptable level (possibly optimality), elicitation costs become too high, or some other termination criterion is met. We propose a generalization of the *current solution (CS)* elicitation strategy, first described in [4]. This strategy has been shown empirically to be very effective in reducing minimax regret with few queries in several domains [4, 5]. The CS strategy considers only parameters involved in defining minimax regret (i.e., the current regret-minimizing solution $\mathbf{x}^*$ and the adversary's witness $\mathbf{x}^w$), and asks a query about the parameter that offers the largest potential reduction in regret. We define below how we *score* various query types, and then define potential query strategies.

**Local queries** The pairwise regret of regret-minimizing outcome $\mathbf{x}^*$ and witness $\mathbf{x}^w$ (the current solution) is:

$$R(\mathbf{x}^*, \mathbf{x}^w, \mathbf{U}) = \max_{\{u_j^\top, u_j^\perp\}} (u_j^\top - u_j^\perp) \sum_j \sum_{i \in 1..N_j} \max_{\{v_j^i\}} C_j^i \, v_j^i,$$

$$= \sum_j (\dot{u}_j^\top - \dot{u}_j^\perp) \sum_{i \in 1..N_j} C_j^i \, \dot{v}_j^i,$$

where $C_j^i = C_{\mathbf{x}_j^w}^i - C_{\mathbf{x}_j^*}^i$, and $\{\dot{u}_j^\top, \dot{u}_j^\perp, \dot{v}_j^i\}$ are utility parameter values that maximize regret. A local bound query adds a constraint on a local parameter $v_j^i$. We wish to find the parameter $v_j^i$ that offers the largest *potential* reduction in the pairwise regret $R(\mathbf{x}^*, \mathbf{x}^w, \mathbf{U})$ at the current solution, hence in the overall minimax regret. The linear constraints on local parameters induce a polytope defining the feasible space for the parameters *for each factor*. Our elicitation strategies use the bounding hyperrectangle of this polytope as an approximation of this feasible region. This allows for quick computation of query quality. (The bounding hyperrectangle can be computed by solving two very small LPs, linear in factor size.) Let $gap_j^i = v_j^i\!\uparrow - v_j^i\!\downarrow$. If we ask a bound query about the midpoint of the gap, the response narrows the gap by half (either lowering the upper bound or raising the lower bound). The impact of constraining $v_j^i$ on the pairwise regret $R(\mathbf{x}^*, \mathbf{x}^w, \mathbf{U})$ is mediated by the magnitude of a constant $C_j^i$ and the current value of a scaling factor $(\dot{u}_j^\top - \dot{u}_j^\perp)$. We define the heuristic *score* for querying parameter $v_j^i$, a measure of its potential for reducing minimax regret, as:

$$S(v_j^i) = (\dot{u}_j^\top - \dot{u}_j^\perp) \, \mathrm{abs}(C_j^i) \, gap_j^i/2$$

The best bound query is that with the highest score. Determining this is linear in the number of GAI parameters.

Scoring local comparison queries is a more complicated, since it is more difficult to estimate the impact of adding a linear constraint on minimax regret. We again approximate the feasible local parameter space with a bounding hyperrectangle. Given the current solution, we consider a list of all pairs $\{(v_j^i, v_j^k)\}$ such that: (a) $C_j^i \neq 0$ and $C_j^k \neq 0$; (b) $v_j^i\!\uparrow \geq v_j^k\!\downarrow$ and $v_j^k\!\uparrow \geq v_j^i\!\downarrow$; and (c) the relationship between $v_j^i$ and $v_j^k$ is not known due to earlier queries. These conditions severely limit the number of pairs one must consider when determining the best local comparison query. The first condition eliminates many parameters from consideration because most coefficients $C_j^i$ are zero. The second checks the bounds for implied relationships. Finally, the relationship between two parameters might already be known beforehand due to prior constraints or transitive closure of previous comparison constraints.

For each pair $(v_j^i, v_j^k)$ considered, we compute a heuristic score as follows. First, we project the bounding hyperrectangle on the plane of the two parameters we are considering; the comparison constraint divides our 2-D rectangle along the 45-degree line. Fig. 1 shows all four cases and demonstrates that, after a response to a comparison query, the values of the parameters $v_j^i, v_j^k$ (as well as the current level of regret) either remain the same, or they are pushed to lie at one of the two intersections of the diagonal with the bounding rectangle. In the latter case, the reduction in local regret can be approximated by

$$r_j^{i,k} = C_j^i \dot{v}_j^i + C_j^k \dot{v}_j^k - \max(C_j^i t_1 + C_j^k t_1, C_j^i t_2 + C_j^k t_2),$$

where $(t_1, t_1)$, $(t_2, t_2)$ are the coordinates of the two intersections. The heuristic score for comparing $\mathbf{x}_j^i$ to $\mathbf{x}_j^k$ is:

$$S(\mathbf{x}_j^i, \mathbf{x}_j^k) = \left(\dot{u}_j^\top - \dot{u}_j^\perp\right) r_j^{i,k}.$$

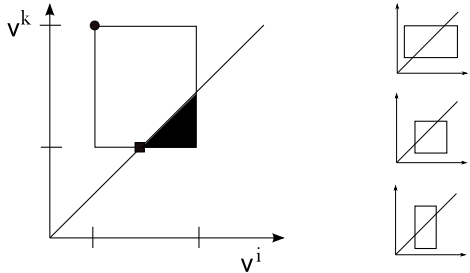

Figure 1: Four different ways to bisect a bounding rectangle. In all cases, if the response to a comparison query eliminates the part of the rectangle which contained the current solution point $(\dot{v}_j^i, \dot{v}_j^k)$ (marked with a circle), the new solution point (marked with a square) is one of the two intersections of the diagonal and the bounding rectangle. The shaded area approximates feasible parameter space after a response to a comparison query.

The complexity of finding the best comparison query is linear in the number factors and quadratic in the number of local outcomes in each factor.

**Global queries** We use similar heuristic methods to compute the score of global anchor queries. In this case, we look at the impact of imposing constraints on anchor parameters $u_j^\top, u_j^\perp$, while keeping local regrets $\sum_i C_j^i \ \dot{v}_j^i$ constant. The resulting heuristic scores for both local and global queries are commensurable, allow comparison of different query types during elicitation.

**Combining different queries** If all types of queries are available, we can simply choose the next query to ask based on the heuristic score $S$ described above. However, in general we want to consider not only the impact of a query in reducing regret, but also its cost to a user. Global queries are generally harder to answer than local queries; similarly, most users will find comparing two outcomes easier than dealing with bound queries (which require some numerical calibration w.r.t. anchors). As such, the scores above are viewed as ranking queries of a *specific type* relative to each other. We can compare queries of different types by scaling these scores by, for example, cost factors that penalize different types of queries.[6]

---

[6] Queries of a single type could also be differentiated by various means (e.g., the number of attributes involved, the number set to non-reference levels, etc.).

Instead, we consider several strategies that combine different query types without explicitly differentiating for cost; but we examine strategies that use only the "easiest" queries. The *LC strategy* uses only local comparison queries; when our heuristic cannot recommend a comparison query, a comparison query is then picked at random. If instead of a random comparison query we select the best local bound query, we get the *LC(LB) strategy*. The *LB strategy* uses only local bound queries. The remaining strategies do not favor any query type, but simply recommend a query from the set of allowed types with the highest score: *LC+LB* combines local comparison and bound queries, and *AB+LC+LB* and *AB+LB* mix global anchor bound queries with local queries.

**Experimental results** We tested our CS elicitation strategies on the car rental configuration problem from [4, 6] and a small apartment rental database problem. The car-rental problem is modeled with 26 attributes that specify various attributes of a car relevant to typical rental decisions. The domain sizes of the attributes range from two to nine values, resulting in $6.1 \times 10^{10}$ possible configurations. The GAI model consists of 13 local factors, each defined on at most five variables; the model has 378 utility parameters. There are ten hard constraints defining feasible configurations. The apartment rental problem comprises a database of 186 apartments, described by eight attributes, each having between two and 33 domain values. The GAI model has five factors, and can be specified with 156 utility parameters. The implementation was in Python, using CPLEX 9.1 to solve MIPs in the car-rental problem (the apartment database requires no MIPs). Computing the regret-minimizing solution, which has to updated after each query, takes about 1 second; determining the next query for any given strategy is even faster. Thus our approach admits real-time interaction.

We evaluated the six query strategies described above. Fig. 2 shows their performance on (a) the car rental configuration problem, and (b) the apartment rental database problem. The results are averaged over 20 random samples of the underlying user utilities as well as random prior bounds on utility parameters. The upper anchor bounds are drawn uniformly from [1,50], and lower bounds from [-50,-1]. The LVF bounds are drawn uniformly from [0,1].

With the exception of the LC strategy, all strategies (including those that use only local queries) exhibit a sharp initial reduction in minimax regret (from .30 to .05 with less than 40 interactions in the car-rental case). This means that in many cases we can either avoid costly global queries altogether or use them only in situations where very strict worst-case loss guarantees are required. Even though the LC strategy does not perform as well as bound query strategies, we note that comparison queries (which are generally less costly in terms of user effort, time and accuracy than bound queries) are very effective during the first ten or so interactions, and do not hinder the performance of strategies in which they are used together with bound queries. Not surprisingly, only strategies that use anchor queries (AB+LC+LB and AB+LB) are able to reduce the regret level to zero; however, the the performance of local-queries-only strategies, such as LC(LB), LC+LB and LB is very encouraging.

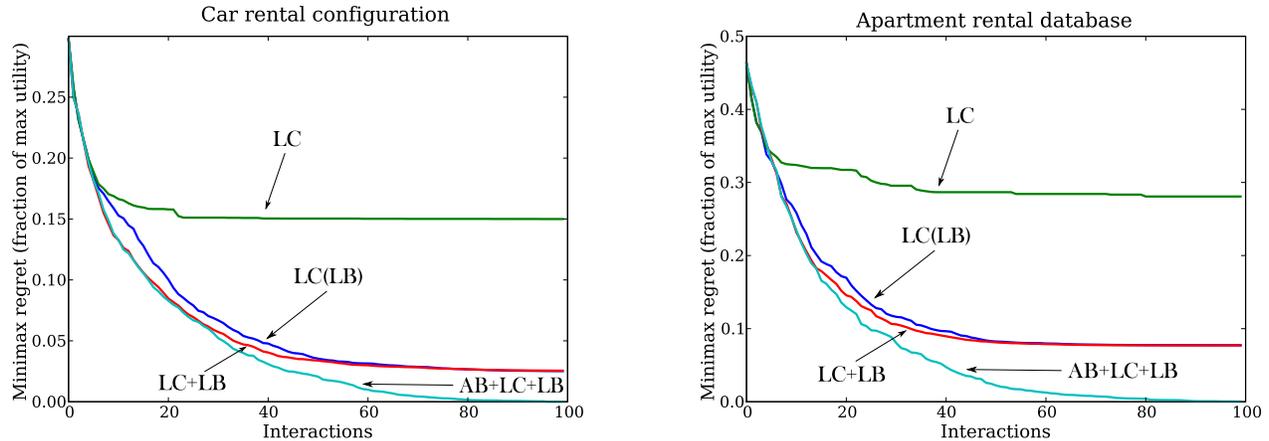

Figure 2: The performance of different query strategies on a) car rental configuration problem; b) apartment rental catalog problem. After averaging over 20 random instantiations of user utilities, the LB strategy curve was virtually indistinguishable from LC+LB; similarly, AB+LB was very close to AB+LC+LB. We omit these two curves for clarity.

## 6 Conclusions

We have provided a semantically justifiable approach to elicitation of utility functions in GAI models using the minimax regret decision criterion. The structure of a GAI model facilitates both elicitation and decision making via the semantically sound separation of local and global components. We described suitable forms of local and global queries and developed techniques for computing minimax optimal decisions under strict utility uncertainty, captured by linear constraints on the parameters of the GAI model. Our elicitation strategies combine both local and global queries and provide a practical way to make good decisions while minimizing user interaction.

We are currently pursuing several extensions of this work. We are: investigating techniques for the effective elicitation of GAI utility structure (something we take as given in this work); exploring the incorporation of probabilistic knowledge of utility parameters to help guide elicitation (while still considering regret in making final decisions [14]); and experimenting with additional query types. Query strategies that take into account explicit query costs are of interest, too. Finally, experiments with human decision makers will allow us to consider the impact of psychological and behavioral issues—such as framing and ordering effects, sensitivity analysis, and different modes of interaction—on our normative model of elicitation.